\documentclass[sigconf]{acmart}

\AtBeginDocument{%
  }

\setcopyright{none}
\copyrightyear{2026}
\acmYear{2026}
\acmConference[DAI '26]{8th International Conference on Distributed Artificial Intelligence}{November 29--December 2, 2026}{Hong Kong}
\acmBooktitle{8th International Conference on Distributed Artificial Intelligence (DAI '26), November 29--December 2, 2026, Hong Kong}
\title{MetaRoute-Bench: Evaluating Meta-Decision Policies for Agentic Workflows}

\author{Natan Vidra}
\affiliation{%
  \institution{Anote AI}
  \city{New York}
  \country{United States}}
\email{nvidra@anote.ai}

\author{Alina Kapanova}
\affiliation{%
  \institution{Anote AI, Cornell University}
  \city{New York}
  \country{United States}}
\email{ak2765@cornell.edu}

\author{Arun Kanhai}
\affiliation{%
  \institution{Anote AI, CUNY}
  \city{New York}
  \country{United States}}
\email{arun.kanhai55@qmail.cuny.edu}

\author{Spurthi Setty}
\affiliation{%
  \institution{Stevens Institute of Technology}
  \city{New York}
  \country{United States}}
\email{ssetty2@stevens.edu}
\renewcommand{\shortauthors}{Vidra et al.}

\begin{document}

\begin{abstract}
Agentic systems must repeatedly decide whether to answer directly, decompose a task, invoke a tool, execute code, delegate to a specialist, verify an intermediate result, or recover from failure. These meta-decisions affect not only task success but also operating cost and latency, yet they are often embedded inside an orchestration framework and evaluated only through aggregate task accuracy. We present MetaRoute-Bench, an open, inspectable framework for comparing meta-decision policies under a shared execution model. The initial benchmark contains 180 synthetic task profiles spanning data analysis, research, and document processing, eight routing policies, and 30 paired random seeds. Across 43,200 traces, a task-aware compositional policy achieves 79.4\% success compared with 76.7\% for a strong workload-specific static policy, 67.4\% for one-shot task routing, and 52.9\% for direct answering. Relative to the static policy, this is a 2.7 percentage-point improvement (paired 95\% CI: $\pm$2.0 points) at 4.7\% higher mean cost and 6.4\% higher latency. Ablations show the largest losses when route composition is restricted to one operation and when verification is removed. These results are generated by a seeded offline execution model rather than a live deployment; accordingly, our primary contribution is a reproducible evaluation method and an analysis of routing-policy tradeoffs, not evidence of production effectiveness. We release task generation, policies, traces, tests, and analysis artifacts to support live-system validation.
\end{abstract}



\maketitle

\section{Introduction}

An agentic application is more than a language model call. It is a distributed decision process involving models, retrieval systems, code executors, specialist agents, verification stages, retry logic, and human escalation. Before any component can help, the system must decide which component to invoke and when. A poor meta-decision can waste time on unnecessary decomposition, call an irrelevant tool, execute unsafe or unhelpful code, or stop before a result has been checked.

Existing research has established the value of interleaving reasoning and action~\cite{yao2023react}, learning API use~\cite{schick2023toolformer,qin2024toolllm}, routing among models~\cite{chen2024frugalgpt,ong2025routellm}, and optimizing function-call plans~\cite{kim2024llmcompiler}. Agent benchmarks reveal persistent failures in long-horizon reasoning and decision making~\cite{liu2024agentbench}, while domain benchmarks increasingly emphasize executable outcomes~\cite{jimenez2024swebench,yao2024taubench}. However, system builders still lack a small, inspectable protocol for isolating the policy that chooses among reasoning modes and measuring its success, cost, and latency tradeoffs.

This paper introduces MetaRoute-Bench, an offline benchmark and trace format for that meta-decision layer. We ask: \emph{under controlled execution assumptions, does task-aware adaptive routing improve operational outcomes over fixed and one-shot policies?} Our contributions are:

\begin{itemize}
  \item a typed framework separating task profiles, routing policies, execution, traces, and evaluation;
  \item a balanced suite of 180 synthetic profiles across three operational workload families;
  \item a paired comparison of eight policies over 30 seeds and 43,200 traces; and
  \item ablations and workload-level analyses identifying when composition, verification, decomposition, and recovery affect outcomes.
\end{itemize}

The implementation, benchmark profiles, experiment scripts, raw traces, and reproducibility materials are available at \url{https://github.com/anote-ai/research-coageneration}.

The benchmark is intentionally transparent, but currently simulated. This boundary is central: the reported values validate the framework's reproducibility and diagnostic behavior, while live models, tools, and organizational workloads are required to establish external validity.

\section{Related Work}

\textbf{Adaptive routing and efficient inference.} Prior routers establish that computational strategy should vary with the request rather than remain fixed. Adaptive-RAG classifies question complexity and selects among no retrieval, single-step retrieval, and iterative retrieval, improving the accuracy--efficiency balance~\cite{jeong2024adaptiverag}. CP-Router similarly uses uncertainty to choose between a standard language model and a longer-reasoning model~\cite{su2026cprouter}. FrugalGPT and RouteLLM route requests among models to balance quality and cost~\cite{chen2024frugalgpt,ong2025routellm}, while ZeroRouter explicitly optimizes accuracy, cost, and latency and supports onboarding unseen models~\cite{yan2026zerorouter}. These systems motivate our task-difficulty thresholds and separate reporting of success, cost, and latency. MetaRoute-Bench differs by routing among workflow operations rather than only retrieval modes or model endpoints; its current threshold policy is transparent and hand specified rather than learned or uncertainty calibrated.

\textbf{Planning, decomposition, and verification.} ReAct interleaves reasoning and environment actions so plans can evolve from observations~\cite{yao2023react}. ACPBench uses formal planning domains to synthesize scalable tasks with provably correct answers and shows that current models retain uneven planning abilities~\cite{kokel2025acpbench}. SPIRAL assigns proposing, simulation, and critique to specialized agents inside grounded reflective search~\cite{zhang2026spiral}. Together, these works support treating decomposition and verification as separable operations and motivate our ablations of each. Our implementation is intentionally less ambitious: it composes an annotated route once and permits an executor-level retry, but does not perform search or observation-conditioned replanning.

\textbf{Tool selection and execution.} Toolformer learns whether, when, and how to invoke APIs~\cite{schick2023toolformer}, and ToolLLM scales tool learning to thousands of real APIs~\cite{qin2024toolllm}. API-Bank separates planning, API retrieval, and API calling in a runnable benchmark~\cite{li2023apibank}; RESTful-Llama demonstrates an industry-oriented path from natural-language requests and API documentation to REST calls~\cite{xu2024restful}. AnyTool adds hierarchical retrieval and reflection~\cite{du2024anytool}, while LLMCompiler optimizes function-call plans for latency and cost~\cite{kim2024llmcompiler}. This literature justifies representing tool use, code execution, delegation, and verification as explicit trace events with independent costs and failures. Unlike API-Bank or RESTful-Llama, the present study simulates those events and therefore cannot establish live tool-call robustness.

\textbf{Agent evaluation and trajectory diagnosis.} AgentBench evaluates agents across interactive environments and identifies long-horizon reasoning and decision-making failures~\cite{liu2024agentbench}. SWE-bench grounds coding-agent evaluation in repository issues~\cite{jimenez2024swebench}, and $\tau$-bench evaluates conversational tool agents under domain policies~\cite{yao2024taubench}. AgentDiagnose argues that final success alone obscures decomposition, observation reading, verification, and backtracking behavior, and instead analyzes full trajectories~\cite{ou2025agentdiagnose}. These findings directly motivate our typed traces, failure taxonomy, action-frequency analysis, and workload slices. MetaRoute-Bench contributes a controlled policy-comparison layer; it complements rather than replaces benchmarks with executable tasks and real environments.

\section{Framework}

\subsection{Design Requirements}

MetaRoute-Bench is designed around four requirements derived from operational agent evaluation. \emph{Policy isolation} requires routing logic to be interchangeable without changing tasks or the executor. \emph{Paired evaluation} requires every policy to encounter the same profiles and seed schedule. \emph{Trace completeness}, consistent with trajectory-level diagnosis~\cite{ou2025agentdiagnose}, requires the artifact to retain actions, failures, retries, cost, and latency rather than only final correctness. Finally, \emph{assumption visibility} requires simulator parameters to remain inspectable and configurable. These requirements make the artifact useful for controlled diagnostics today and for later shadow evaluation against live systems.

The unit of evaluation is a route, not an isolated model response. This distinction matters because two systems can use the same base model yet produce different operational behavior through decomposition, tool access, verification, and retry policies. Conversely, a more capable model can be operationally inferior if its controller invokes costly components unnecessarily. MetaRoute-Bench therefore treats the controller as an independently testable system component.

\subsection{Meta-Decision Process}

Each task $x$ has a workload, difficulty, ambiguity, operation-need annotations, and cost and latency budgets. The present policies map task metadata to a complete route $r$:

\begin{equation}
r = \pi(x), \quad r \subseteq \{D,T,C,G,V,A\},
\end{equation}

where $D$ denotes decomposition, $T$ tool use, $C$ code execution, $G$ delegation, $V$ verification, and $A$ final answering. Every route must end in $A$; operations are unique in the current implementation. A trace records actions, success, expected success, cost, latency, retries, budget compliance, confidence, and failure mode. The framework can be extended to policies over execution history, but the evaluated policy adapts only through one executor-level retry after a failed operation; it does not replan from arbitrary intermediate outputs.

We report success, cost, and latency separately. For ranking in the command-line summary only, we define a secondary utility

\begin{equation}
U = S - 0.025K - 0.0015L,
\end{equation}

where $S$ is success rate, $K$ mean normalized cost, and $L$ mean latency in seconds. Conclusions do not depend solely on this weighting.

\subsection{System Architecture}

The implementation separates five interfaces: (1) a deterministic workload generator; (2) routing policies; (3) a seeded offline executor; (4) typed trace export; and (5) aggregate and paired evaluation. This separation permits replacement of the simulator with live model and tool adapters without changing policy or analysis interfaces.

\begin{figure}[t]
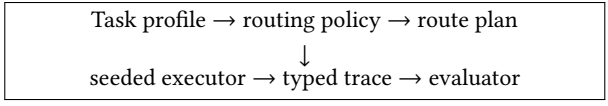

  \centering
  \fbox{\begin{minipage}{0.91\columnwidth}
    \centering
    Task profile $\rightarrow$ routing policy $\rightarrow$ route plan\\[2pt]
    $\downarrow$\\[-2pt]
    seeded executor $\rightarrow$ typed trace $\rightarrow$ evaluator
  \end{minipage}}
  \caption{MetaRoute-Bench separates task generation, route selection, execution, and evaluation. Live adapters can replace the seeded executor while preserving trace and metric interfaces.}
  \Description{A pipeline begins with a task profile, passes through a routing policy to form a route plan, executes the plan, records a typed trace, and sends the trace to an evaluator.}
  \label{fig:architecture}
\end{figure}

The executor makes its assumptions explicit. Each operation has a normalized cost, latency, benefit scaled by task need, and---for external tools, code, and delegation---a failure probability. Difficulty and ambiguity reduce base success. Unnecessary operations impose overhead. Adaptive recovery makes one retry after an execution failure. All randomness is keyed by seed, task identifier, and policy name.

\subsection{Policies}

We evaluate four fixed policies (direct, always decompose, always tool, and always code), a random route policy, a workload-specific static rule table, a one-shot router selecting the highest annotated operation need, and the proposed task-aware compositional policy, labeled \emph{adaptive} in the artifacts. It selects up to three operations above a difficulty-dependent threshold, orders decomposition first and verification last, and enables one recovery attempt. Both one-shot and adaptive policies receive the same task-level need annotations; their comparison therefore isolates route composition and recovery rather than feature prediction.

The adaptive policy uses a threshold of .66 for difficulty levels one and two and .56 for levels three and four. This coarse complexity conditioning follows the same design principle as Adaptive-RAG~\cite{jeong2024adaptiverag}, although our thresholds are fixed rather than classifier learned. If no operation clears the threshold, the policy selects the highest-scoring operation. A maximum of three support operations prevents unbounded orchestration overhead. Decomposition is moved to the beginning of a route because it affects downstream work allocation, while verification is moved to the end because it evaluates the assembled result. The current policy is deliberately transparent: every decision can be reconstructed from exported task annotations and constants. This favors auditability over model flexibility and provides a reproducible baseline for future learned or uncertainty-aware routers~\cite{su2026cprouter,yan2026zerorouter}.

The static workload baseline is intentionally strong. Data-analysis profiles execute code and verify, research profiles decompose and use a tool, and document-processing profiles use a tool and verify. It represents the kind of rule table an engineering team might deploy before investing in a learned or task-aware controller. The one-shot policy receives richer task annotations but can choose only one support operation. Comparing these policies distinguishes three questions: whether orchestration helps at all, whether workload rules are sufficient, and whether composing multiple task-specific operations adds value.

\section{Evaluation}

\subsection{Workloads and Protocol}

The suite contains 60 task profiles for each of data analysis, research, and document processing. Profiles cover four difficulty levels and use seeded variation around workload-specific needs. Data analysis emphasizes code and verification, research emphasizes decomposition and tools, and document processing emphasizes tools and verification. These profiles represent workflow characteristics rather than natural-language task instances.

\begin{table}[t]
\caption{Mean annotations for the 180 task profiles. Dcmp. denotes decomposition and Verif. denotes verification.}
\label{tab:workloads}
\small
\begin{tabular}{lrrrrrr}
\toprule
Workload & Amb. & Dcmp. & Tool & Code & Deleg. & Verif. \\
\midrule
Data analysis & .40 & .50 & .35 & .76 & .32 & .64 \\
Research & .39 & .67 & .78 & .19 & .56 & .67 \\
Document proc. & .38 & .39 & .65 & .46 & .25 & .74 \\
\bottomrule
\end{tabular}
\end{table}

Table~\ref{tab:workloads} summarizes the resulting suite. The profiles are balanced by workload and difficulty, but intentionally heterogeneous within each workload: every need receives uniform jitter of up to .25 before clipping. This prevents the workload label from fully determining the best route and creates cases where a workload-level rule is unnecessarily expensive or omits a useful operation. Scalable synthetic evaluation has precedent in planning benchmarks such as ACPBench~\cite{kokel2025acpbench}; however, our annotations are not backed by formal semantics or provably correct plans. They should be viewed as controlled routing signals rather than a realistic task-understanding stage.

We execute every policy on every profile for 30 paired seeds. This produces 5,400 traces per policy and 43,200 main-experiment traces. We report mean success and a 95\% confidence interval computed over seed-level success rates, plus mean normalized cost, latency, and cost per success. Paired differences use the same seed-level aggregation. The complete trace file is exported as CSV.

\subsection{Main Results}

\begin{table}[t]
\caption{Main benchmark results. CI is the 95\% interval half-width for success over 30 seeds.}
\label{tab:main}
\small
\begin{tabular}{lrrrr}
\toprule
Policy & Success & CI & Cost & Lat. (s) \\
\midrule
Adaptive & \textbf{.794} & .012 & 2.91 & 20.45 \\
Static workload & .767 & .013 & 2.78 & 19.23 \\
One-shot & .674 & .012 & 1.83 & 12.46 \\
Always tool & .637 & .011 & 1.96 & 13.44 \\
Random & .635 & .012 & 2.54 & 17.86 \\
Always decompose & .610 & .013 & 1.34 & 8.40 \\
Always code & .608 & .011 & 2.24 & 15.68 \\
Direct & .529 & .012 & .73 & 4.48 \\
\bottomrule
\end{tabular}
\end{table}

Table~\ref{tab:main} shows a clear operational tradeoff. Adaptive routing has the highest success rate, but direct answering is least expensive and fastest. Against the strongest baseline, static workload routing, adaptive routing improves success by 2.74 points (paired 95\% CI $\pm$1.96), while increasing cost by 0.13 units and latency by 1.22 seconds. Compared with one-shot routing, adaptive routing improves success by 12.02 points ($\pm$1.87), with 1.09 additional cost units and 8.00 additional seconds.

\begin{figure*}[t]
  \centering
  \includegraphics[width=0.92\textwidth]{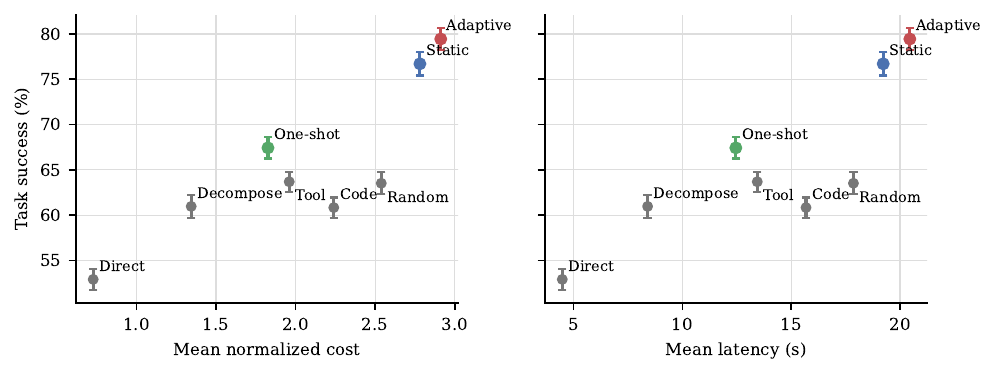}
  \caption{Success versus mean normalized cost (left) and mean latency (right). Error bars show 95\% confidence intervals over seed-level success. No policy dominates all three dimensions: direct answering is cheapest and fastest, while adaptive routing has the highest success.}
  \Description{Two scatter plots compare eight routing policies. In both plots, adaptive routing has the highest success and highest cost or latency, static workload routing is second in success and slightly cheaper, and direct answering is cheapest and fastest but has the lowest success.}
  \label{fig:tradeoffs}
\end{figure*}

Figure~\ref{fig:tradeoffs} makes the absence of a single universally best policy explicit. Direct answering and always-decompose are attractive when latency or cost dominates, while adaptive and static routing occupy the high-success region. The adaptive policy's cost per successful task is 3.67 units, compared with 3.63 for static routing and 2.71 for one-shot routing. Thus, its success advantage over static routing does not translate into lower cost per success under the current weights. A deployment should select a policy from this frontier using service-level objectives rather than ranking by success alone.

Workload-level adaptive success is .802 for data analysis, .835 for research, and .746 for document processing. The static policy is particularly competitive for data analysis (.791), but trails more on research (.786 versus .835). This suggests that task-level composition has the most value where decomposition and information access interact.

\subsection{Ablations}

\begin{table}[t]
\caption{Adaptive-policy ablations over 5,400 traces each.}
\label{tab:ablation}
\small
\begin{tabular}{lrrr}
\toprule
Policy & Success & Cost & Lat. (s) \\
\midrule
Adaptive & \textbf{.794} & 2.91 & 20.45 \\
No recovery & .783 & 2.86 & 19.98 \\
No decomposition & .769 & 2.88 & 20.44 \\
No verification & .747 & 2.69 & 18.96 \\
Single operation & .681 & 1.85 & 12.68 \\
\bottomrule
\end{tabular}
\end{table}

Restricting routes to one operation produces the largest reduction: 11.30 points ($\pm$1.85). Removing verification reduces success by 4.70 points ($\pm$1.68), and removing decomposition reduces it by 2.57 points ($\pm$1.68). The 1.13-point recovery difference has an interval of $\pm$1.70 and therefore does not support a confident recovery benefit in this experiment. This is a useful negative finding: recovery is operationally plausible, but the current failure frequency and sample design do not establish its effect.

\subsection{Routing and Failure Analysis}

The adaptive policy verifies 68.3\% of routes, uses a tool in 50.6\%, executes code in 33.3\%, decomposes 28.3\%, and delegates 13.3\%. In contrast, the static policy applies exactly two support operations to every task. The adaptive policy therefore does not improve by simply constructing longer routes; it reallocates operations according to profile-level signals and uses three operations only where multiple needs clear the threshold.

Failure traces separate unrecovered execution failures from ordinary task failures. Adaptive routing records 109 unrecovered tool, code, or delegation failures among 5,400 traces (2.0\%), compared with 408 tool or code failures for static routing (7.6\%). Ordinary task failures are similar: 1,060 for adaptive and 1,050 for static. The lower unrecovered-execution count is consistent with retry behavior, but the no-recovery ablation's confidence interval includes zero. We therefore treat this pattern as diagnostic evidence about the trace model, not proof that recovery improves overall success.

Document processing is the weakest adaptive workload at .746 success. Its high verification and tool needs make routes vulnerable to external-operation failure while offering less benefit from decomposition. Research shows the largest advantage over static routing: 4.89 points. These differences illustrate why aggregate scores should be accompanied by workload slices before an orchestration policy is deployed broadly.

\section{Industry Application and Lessons}

First, a strong static policy deserves inclusion in agent evaluations. It captures much of the benefit of adaptive routing and is simpler to inspect and operate. Second, success improvements must be reported alongside cost and latency: the adaptive policy is neither cheapest nor fastest. Third, route composition matters more than any single fixed action in this model. Fourth, structured traces make policy behavior auditable; aggregate accuracy alone cannot reveal unnecessary calls, retry behavior, or workload-specific regressions.

\subsection{Integration Pattern}

For deployment, the framework should sit above existing model and tool adapters. A task-intake service would provide observable metadata and available capabilities; the router would return a structured route plan; an executor would enforce permissions and budgets; and the trace service would record decisions, outcomes, and stop reasons. This architecture does not require storage of private chain-of-thought. Short structured rationales, confidence values, tool inputs and outputs, timing, and failure codes are sufficient for policy analysis.

The separation between router and executor is operationally important. The router may recommend code execution or delegation, but the executor remains responsible for sandboxing, access control, timeouts, data residency, and allowlists. Hard controls should not depend on the router's language-model judgment. Likewise, verification should use an independent check where possible rather than asking the same component to approve its own output.

\subsection{Deployment Protocol}

A production evaluation should proceed in four stages. First, replay historical, consented traces offline and compare proposed routes with the incumbent policy. Second, run in shadow mode: generate route decisions without executing them, then estimate disagreement, expected cost, and policy coverage. Third, execute only low-risk tasks under hard cost and latency limits, retaining a static fallback and human escalation. Finally, use a randomized or stepped-wedge comparison where organizational constraints permit, measuring end-to-end task success rather than proxy judgments alone.

Operational acceptance criteria should be specified before testing. Examples include a lower bound on task success, a maximum p95 latency, a cost-per-success ceiling, and a maximum unrecovered tool-failure rate. Results should also be sliced by workload, difficulty, data sensitivity, and route type. A global average can hide a policy that is beneficial for research tasks but harmful for document workflows.

\subsection{Practical Lessons}

The study yields four immediate lessons. First, strong static policies are credible production baselines, not strawmen. Second, route composition can improve success, but each additional operation consumes budget and expands the failure surface. Third, verification appears valuable in the simulator and should be isolated in live ablations rather than assumed beneficial. Fourth, complete traces turn routing into an observable engineering problem: teams can inspect unnecessary calls, failure recovery, policy disagreement, and workload regressions instead of debugging from final responses alone.

\section{Limitations, Ethics, and Next Steps}

The primary limitation is external validity. Outcomes are sampled from a hand-specified execution model whose operation benefits depend on the same need dimensions supplied to task-aware policies. The adaptive policy should therefore be interpreted as an annotated upper-bound controller, not a learned router. Cost units and latency values are normalized assumptions, not invoices or wall-clock measurements. Task profiles do not contain natural-language inputs, live APIs, concurrent agents, security constraints, or human judgments.

Construct validity is also limited. ``Success'' is a Bernoulli outcome generated from the simulator rather than an independently graded artifact, and the confidence intervals quantify sampling variation under the fixed model, not uncertainty about the model assumptions themselves. The utility weights are illustrative and can change policy rankings. Internal validity is stronger because policies share task profiles, seeds, budgets, and executor logic, but policy-specific random streams mean traces are paired at the seed aggregate rather than by identical random draws for every operation.

The generated workload profiles encode our expectations about which operations help each task family. This makes the benchmark appropriate for checking whether evaluation machinery behaves coherently, but it also creates a form of evaluator-policy alignment. A live study must derive routing signals from task text or operational metadata without revealing ground-truth operation utilities. It should also test distribution shift, missing tools, correlated failures, concurrent execution, and adversarial or malformed inputs.

These limitations prevent claims of deployment readiness. The next evaluation must replace task annotations with predictions derived from task text, replay real anonymized traces, and then run controlled live tools under fixed budgets. A learned router, model-routing baselines, calibration analysis, and human escalation should also be added. We publish all simulator constants and raw traces so these assumptions can be challenged rather than hidden.

Agent routing can create privacy, security, and accountability risks when tasks are delegated to external services or code is executed. Practical deployments need data-minimization rules, sandboxed execution, allowlisted tools, access control, audit logs, and human review for high-impact actions. The current benchmark performs no real external action and contains no personal or proprietary data.

\section{Conclusion}

MetaRoute-Bench isolates the orchestration policy governing how an agentic system decomposes, routes, executes, verifies, and recovers. In a reproducible offline study, task-aware adaptive composition improves success over fixed and one-shot policies while incurring measurable cost and latency. The framework and traces provide a foundation for the more consequential next step: validation on live, production-representative workflows.

\begin{acks}
This work was conducted with organizational support from Anote AI. We thank the Anote AI research and engineering team for research infrastructure, technical feedback, and support in developing the reproducible evaluation artifact. No human participants or proprietary data were used in the reported offline benchmark.
\end{acks}

\section*{Generative AI Disclosure}
OpenAI Codex materially assisted with software implementation, experiment scripting, literature organization, and manuscript drafting. The authors are responsible for reviewing the code, verifying the generated results against the released artifacts, checking all citations, and approving the final claims and text.

\bibliographystyle{ACM-Reference-Format}
\bibliography{references}

\appendix

\section{Execution Model}

The offline executor uses explicit per-operation assumptions, listed in Table~\ref{tab:parameters}. Difficulty multiplies cost and latency by $1+0.08(d-1)$. Base success is $0.69-0.075(d-1)-0.13a$, where $d$ is difficulty and $a$ is ambiguity. Each selected operation adds a benefit proportional to its annotated task need and a small penalty when unnecessary. Routes longer than two support operations incur coordination overhead. External operations can fail; the adaptive condition permits one retry with additional cost and latency. Probabilities are clipped to $[0.02,0.98]$. The implementation in \texttt{src/metarouter/simulator.py} is authoritative.

\begin{table}[h]
\caption{Offline executor parameters. Failure probability is zero where omitted.}
\label{tab:parameters}
\small
\begin{tabular}{lrrr}
\toprule
Operation & Cost & Latency (s) & Failure \\
\midrule
Decompose & .55 & 3.5 & -- \\
Use tool & 1.10 & 8.0 & .08 \\
Execute code & 1.35 & 10.0 & .07 \\
Delegate & 1.65 & 13.0 & .10 \\
Verify & .70 & 5.0 & -- \\
Answer & .65 & 4.0 & -- \\
\bottomrule
\end{tabular}
\end{table}

\section{Reproducibility}

The artifact requires Python 3.10 or later. The following commands regenerate the main and ablation results and verify that the headline manuscript values match the exported CSV files.

\begin{verbatim}
python -m venv .venv
source .venv/bin/activate
pip install -e ".[dev]"
metarouter-benchmark --seeds 30 \
  --output results/meta-routing/dai2026/main
python experiments/meta-routing/dai2026/run_ablations.py
python experiments/meta-routing/dai2026/check_paper_results.py
pytest -q
\end{verbatim}

The artifact exports task profiles, raw traces, policy summaries, paired comparisons, workload-level success, action counts, and failure distributions. The random stream is keyed by seed, task identifier, and policy name.

\end{document}